 \documentclass[sigconf]{acmart}
\usepackage{multirow}
\usepackage{pifont}
\usepackage{microtype}
\AtBeginDocument{%
  }

\copyrightyear{2026}
\acmYear{2026}
\setcopyright{cc}
\setcctype{by}
\acmConference[ICMR '26]{International Conference on Multimedia Retrieval}{June 16--19, 2026}{Amsterdam, Netherlands}
\acmBooktitle{International Conference on Multimedia Retrieval (ICMR '26), June 16--19, 2026, Amsterdam, Netherlands}
\acmDOI{10.1145/3805622.3810735}
\acmISBN{979-8-4007-2617-0/2026/06}

\begin{document}

\title{UAU-Net: Uncertainty-aware Representation Learning and Evidential Classification for Facial Action Unit Detection}

\author{Yuze Li}
\orcid{0009-0002-7012-4749}
\affiliation{%
 \department{School of Artificial Intelligence}
  \institution{Tianjin University}
  \city{Tianjin}
  \country{China}
}
\email{2024244204@tju.edu.cn}

\author{Zhilei Liu}
\orcid{0000-0003-1447-6256}
\authornote{Corresponding author.}
\affiliation{%
 \department{School of Artificial Intelligence}
  \institution{Tianjin University}
  \city{Tianjin}
  \country{China}
}
\email{zhileiliu@tju.edu.cn}

\renewcommand{\shortauthors}{Yuze Li and Zhilei Liu}

\begin{abstract}
Facial action unit (AU) detection remains challenging because it involves heterogeneous, AU-specific uncertainties arising at both the representation and decision stages. Recent methods have improved discriminative feature learning, but they often treat the AU representations as deterministic, overlooking uncertainty caused by visual noise, subject-dependent appearance variations, and ambiguous inter-AU relationships, all of which can substantially degrade robustness. Meanwhile, conventional point-estimation classifiers often provide poorly calibrated confidence, producing overconfident predictions, especially under the severe label imbalance typical of AU datasets. We propose \emph{UAU-Net}, an \textbf{U}ncertainty-aware \textbf{AU} detection framework that explicitly models uncertainty at both stages. At the representation stage, we introduce \emph{CV-AFE}, a conditional VAE (CVAE)-based AU feature extraction module that learns \emph{probabilistic} AU representations by jointly estimating feature means and variances across multiple spatio-temporal scales; conditioning on AU labels further enables CV-AFE to capture uncertainty associated with inter-AU dependencies. At the decision stage, we design \emph{AB-ENN}, an Asymmetric Beta Evidential Neural Network for multi-label AU detection, which parameterizes predictive uncertainty with Beta distributions and mitigates overconfidence via an asymmetric loss tailored to highly imbalanced binary labels. Extensive experiments on BP4D and DISFA show that UAU-Net achieves strong AU detection performance, and further analyses indicate that modeling uncertainty in both representation learning and evidential prediction improves robustness and reliability.

\end{abstract}

%\begin{CCSXML}
%<ccs2012>
%  <concept_id>00000000.0000000.0000000</concept_id>
%  <concept_desc>Do Not Use This Code, Generate the Correct Terms for Your Paper</concept_desc>
%  <concept_significance>500</concept_significance>

%\end{CCSXML}

%\ccsdesc[500]{Do Not Use This Code~Generate the Correct Terms for Your Paper}



%  \includegraphics[width=\textwidth]{figures/TeaserFigure.jpg}
%  \caption{Motivation and overview of the proposed Uncertainty-aware AU detection framework (UAU-Net)}
%  \Description{Different facial action units (AUs) exhibit heterogeneous and AU-specific uncertainty across spatial–temporal feature scales. To address this challenge, we propose a CVAE-based probabilistic representation learning module to explicitly model feature uncertainty, followed by an evidential AU classifier that captures predictive uncertainty. This design leads to more robust and reliable AU detection.}


\maketitle

\section{Introduction}
\label{sec:intro}
Facial action unit (AU) detection recognizes fine-grained facial muscle activations defined by the Facial Action Coding System (FACS)~\cite{ekman1978facial} and is a core problem in affective computing and human-centered multimedia understanding~\cite{li2022deep}. Compared to categorical recognition, AU detection involves reasoning about small, localized changes and is therefore sensitive to nuisance factors such as motion blur, occlusion, changes in illumination, and differences between subjects. These factors introduce substantial~\emph{uncertainty} into both learned representations and final predictions, resulting in ambiguous AU cues and poorly calibrated confidence estimates. Most AU detectors can be abstracted as two sequential stages: (i) AU representation learning/feature extraction—often coupled with explicit inter-AU relationship modeling—and (ii) AU prediction with a multi-label classifier. Uncertainty is intrinsic to both stages. Despite progress, methods remain vulnerable under challenging conditions because they are typically trained as~\emph{point-estimation} pipelines that do not model heterogeneous uncertainty~\cite{li2021your,song2021uncertain}.

\paragraph{\textbf{Representation uncertainty}} Early AU detectors relied on hand-crafted cues such as landmarks or region-based pooling to obtain static AU features~\cite{shao2021jaa}. Recent approaches improve discriminability by modeling contextual dependencies among AUs~\cite{icasspSongL23,niu2019local,shao2019facial} using transformers~\cite{jacob2021facial,yang2023fan} or graph-based models~\cite{song2021uncertain, ijcai2022p173}, and by incorporating temporal dynamics across frames~\cite{li2021integrating}. However, inter-AU reasoning can be unreliable when evidence is weak or conflicting, and most methods do not \emph{explicitly quantify} the uncertainty introduced by relationship modeling. In addition, although VAE-based formulations can capture data uncertainty~\cite{zhu2025towards}, they are often tailored to a single source of uncertainty and do not account for uncertainty amplified by inter-AU dependency modeling.

%These early works overlooked the influence of uncertainty on the robustness of AU detection models.As a consequence, these models exhibit limited capability in dealing with highly uncertain samples. Since AUs are mutually dependent ~\cite{tong2007facial}, Many approaches ~\cite{niu2019local,shao2019facial} attempt to capture contextual relationships among different AUs from the entire facial image, with recent methods primarily relying on transformer-based architectures ~\cite{jacob2021facial,yang2023fan} and graph-based modeling ~\cite{luo2022learning,song2021uncertain}. Some studies also attempt to model temporal dynamics by capturing AU motion patterns across frames ~\cite{li2021integrating,li2021sat,shao2020spatio}. However, uncertainty inevitably arises when modeling the relationships among action units. Although existing AU relationship modeling methods aim to capture inter-AU dependencies and implicitly reduce uncertainty through structured modeling, they do not explicitly quantify the uncertainty arising from the relationships. Moreover, these approaches generally lack a fine-grained analysis of data uncertainty, particularly with respect to how noisy samples and ambiguous AU labels affect the learning process. Other prior studies ~\cite{zhu2025towards} employ VAE-based models to capture data uncertainty. However, the uncertainty they model is limited to a single source and does not account for the additional uncertainty introduced during AU relationship modeling.

\paragraph{\textbf{Predictive uncertainty and label imbalance}} Uncertainty also manifests at the prediction stage~\cite{wang2021can}. Most AU detectors are trained with cross-entropy and produce deterministic probabilities~\cite{li2018eac,shao2021jaa}, which often yields overconfident outputs under distribution shifts or low-quality inputs. Although multi-branch architectures can partially alleviate overconfidence~\cite{shao2026constrained}, they still do not provide an explicit probabilistic characterization of predictive uncertainty. This limitation is further amplified by the severe class imbalance of AU datasets, where negative labels dominate. Existing solutions typically rely on manually tuned re-weighting~\cite{shao2025facial, yan2022weakly}, and some recent work adjusts weights based on uncertainty from relationship learning~\cite{song2021uncertain}. However, such uncertainty primarily reflects inter-AU reasoning rather than the classifier's predictive uncertainty, leaving its role in both training and inference insufficiently explored.

To address these limitations, we propose \textbf{UAU-Net}, an \textbf{U}ncertainty-aware \textbf{AU} detection framework that jointly models \emph{representation uncertainty} and \emph{predictive uncertainty}. At the representation stage, we introduce \emph{CV-AFE}, a conditional variational autoencoder (CVAE)-based AU feature extractor that learns \emph{probabilistic} AU representations by estimating feature means and variances across multiple spatio--temporal scales. By conditioning the latent distribution on AU labels, CV-AFE captures AU-specific uncertainty as well as uncertainty induced by inter-AU dependencies. At the prediction stage, we formulate AU detection as a multi-label binary task and propose an \emph{Asymmetric Beta Evidential Neural Network} (AB-ENN). AB-ENN models each AU's occurrence probability with a Beta distribution, enabling joint estimation of likelihood and predictive uncertainty via subjective opinions rather than deterministic probabilities. To handle extreme positive--negative imbalance, we further derive an asymmetric evidential loss that down-weights easy negatives while emphasizing positives.

%At the feature representation level, we introduce a Conditional Variational Autoencoder (CVAE)–based AU feature extraction module that learns probabilistic AU representations by modeling both feature means and variances across multiple spatial–temporal scales. By conditioning the latent distribution on AU labels, the proposed module effectively captures AU-specific uncertainty as well as inter-AU dependencies, enabling more reliable uncertainty-aware feature fusion.
%At the classification level, we formulate AU detection as a multi formulate AU detection as a multi-label binary task and propose the Asymmetric Beta Evidential Neural Network (AB-ENN). Leveraging Beta distributions, AB-ENN jointly estimates the likelihood and predictive uncertainty of each AU, and infers subjective opinions instead of deterministic probabilities via evidential theory and subjective logic—effectively mitigating overconfidence of traditional point-estimation models. Furthermore, given that EBCE loss, despite its theoretical advantages in evidential learning, suffers from severe positive–negative imbalance in AU detection, we adopt an asymmetric loss formulation that down-weights easy negative samples and emphasizes positive ones to alleviate this issue.

The major contributions of this work are summarized as follows:
\begin{itemize}
    \item We propose \emph{CV-AFE}, a CVAE-based AU feature extraction module that introduces explicit uncertainty modeling into AU representations. By jointly learning feature means and variances across multiple spatio-temporal scales, CV-AFE captures both data uncertainty (e.g., visual noise) and relational uncertainty induced by label-conditioned inter-AU dependency modeling.
    
    \item We introduce \emph{AB-ENN}, an Asymmetric Beta Evidential Neural Network for AU prediction, which models AU occurrence with Beta distributions instead of conventional point estimates. This formulation enables principled predictive uncertainty estimation and alleviates overconfident predictions. We further derive an asymmetric Beta loss to address severe class imbalance by emphasizing positive AU instances while down-weighting easy negatives.
    
    \item Extensive experiments on BP4D~\cite{zhang2014bp4d} and DISFA~\cite{mavadati2013disfa} show that UAU-Net achieves competitive performance, and that explicitly modeling uncertainty at both the representation and prediction stages yields consistent improvements.
    %Experimental results show that our model achieves new state-of-the-art performance on widely used AU benchmark datasets, including BP4D~\cite{zhang2014bp4d} and DISFA~\cite{mavadati2013disfa}. The uncertainty modeling introduced by CV-AFE and AB-ENN at different stages jointly contributes to the overall performance improvement.
\end{itemize}

\begin{figure*}[htbp]
    \centering
    
    \includegraphics[width=\textwidth]{figures/framework-260106.pdf}
    \caption{Architecture of our UAU-Net framework, in which CVAE-based feature extraction (CV-AFE) and asymmetric Beta-ENN (AB-ENN) are the core modules for AU detection. After spatio-temporal global feature extraction, the CV-AFE module learns uncertainty-aware features for each AU from local facial regions. With inter-AU dependencies across regions modeled, AB-ENN formulates AU prediction as a binary evidential learning problem, explicitly capturing predictive uncertainty for robust AU classification. $\otimes$ and $\oplus$ denote element-wise multiplication and addition, respectively; $C$, $S$, and $E$ in circles indicate concatenation, split, and expansion.}
    \Description{Overview of UAU-Net. A spatio-temporal backbone extracts facial features from neighboring frames, CV-AFE converts local AU features into probabilistic embeddings with estimated means and variances, a global module models inter-AU dependencies, and AB-ENN predicts each AU with evidential Beta distributions and uncertainty-aware classification.}
    \label{fig:framework}
\end{figure*}

\section{Related Work}
This section reviews work most relevant to our framework, focusing on facial AU detection and uncertainty-aware learning. We summarize how existing methods learn AU representations and model inter-AU dependencies, and highlight their limitations in explicitly quantifying uncertainty at both the representation and prediction stages.
%We review the previous techniques that are closely relevant to our proposed framework, including facial action unit detection and uncertainty modeling.

\subsection{Facial Action Unit Detection}

Facial AU detection pipelines typically consist of representation learning and AU prediction, and can be broadly categorized into static image-based and spatio-temporal methods~\cite{adyapady2023comprehensive}. Early static approaches improved AU feature quality by exploiting global context and local facial regions with attention mechanisms. For instance, Shao et al.~\cite{shao2019facial} employed adaptive channel-wise and spatial attention to focus on AU-relevant cues. Given that AUs are inherently interdependent~\cite{zhang2018classifier}, many works further incorporate explicit relationship modeling. LP-Net~\cite{niu2019local} leverages sequential models to capture AU dependencies, while Jacob et al.~\cite{jacob2021facial} proposed a transformer-style AU correlation network. Graph-based methods~\cite{li2019semantic,ijcai2022p173} treat AUs as nodes and encode pairwise dependencies as edges, enabling structured reasoning over AU relations.

%Many existing AU detection methods implicitly mitigate various sources of uncertainty on the feature representation. For example, to incorporate global context, several approaches learn AU features from the entire face image using attention mechanisms. Shao et al. ~\cite{shao2019facial} employed adaptive channel-wise and spatial attention to focus on AU-relevant local features. Since AUs are interdependent ~\cite{zhang2018classifier}, many works explicitly model their relationships. LP-Net ~\cite{niu2019local} uses LSTMs to capture sequential AU dependencies, and Jacob et al. ~\cite{jacob2021facial} proposed a transformer-style AU correlation network. Graph-based methods ~\cite{li2019semantic,liu2019relation,luo2022learning} represent AUs as nodes and their pairwise dependencies as edges, allowing explicit relationship modeling. Shao et al. ~\cite{shao2025facial} leverages both global and local attention networks, whose outputs are jointly exploited to reduce the uncertainty encountered in AU classification.

To capture temporal dynamics, recent methods combine spatial relationship modeling with temporal aggregation. HSTR-Net~\cite{song2022heterogeneous} constructs frame-level AU graphs and models AU evolution across frames. Transformer-based approaches~\cite{ijcnnCaoLZ22,li2023knowledge} similarly capture inter-AU dependencies and temporal context via multi-head attention. Yang et al.~\cite{yang2023toward} introduced a Temporal Difference Network (TDN) to extract facial dynamics at specific spatial scales, and MDHR~\cite{wang2024multi} further hierarchically models both local and cross-regional spatio-temporal relationships among AUs for robust occurrence recognition. Although these designs improve discriminability, relationship encoding is typically deterministic, and the uncertainty introduced by inter-AU reasoning is rarely quantified explicitly.

%More recent approaches combine spatial and temporal modeling. Spatio-temporal GNNs ~\cite{shao2020spatio} and Heterogeneous Spatio-temporal Relation Networks (HSTR-Net) ~\cite{song2022heterogeneous} first construct frame-level spatial AU graphs and then model each AU’s temporal evolution using node features across frames. Similarly, transformer-based models ~\cite{li2023knowledge} capture both inter-AU spatial dependencies and temporal inter-frame contexts via multi-head attention. Recently, Yang et al. ~\cite{yang2023toward} introduced a Temporal Difference Network (TDN) to extract facial dynamics at specific spatial scales. MDHR ~\cite{wang2024multi} captures multi-scale facial dynamics and hierarchically models both local and cross-regional spatio-temporal relationships among AUs for robust occurrence recognition. These methods exploit temporal and spatial information of AUs to alleviate the uncertainty of AU relationships during feature extraction. However, they neither explicitly quantify nor analyze such uncertainties, despite the fact that they are crucial for the AU detection task.
 
Recently, several studies have begun to explicitly quantify uncertainty and have reported promising improvements for AU detection. Song et al.~\cite{song2021uncertain} proposed an uncertainty-aware graph neural network with a probabilistic masking mechanism to model and reweight AU dependencies. Zhu et al.~\cite{zhu2025towards} learned stochastic representations for each sample and estimated feature variance via uncertainty-aware score assignment. UA-FER~\cite{zhou2025ua} modeled and rectified uncertainty in visual--textual relationships using evidential deep learning (EDL)~\cite{sensoy2018evidential}, improving robustness and reliability. Chen et al.~\cite{chen2022uncertainty} introduced data uncertainty learning together with a context-aware contrastive objective for robust pedestrian representation learning.

Nevertheless, most existing methods focus on a \emph{single} uncertainty source (e.g., feature noise \emph{or} output uncertainty) and do not jointly account for uncertainty introduced by inter-AU reasoning \emph{and} the highly imbalanced multi-label prediction setting. In contrast, our framework provides a unified uncertainty-aware design that covers both representation learning and evidential decision making for AU detection.

\subsection{Uncertainty Modeling}

Uncertainty modeling or uncertainty-aware learning aims to quantify prediction reliability, and is commonly discussed in terms of data (aleatoric) uncertainty and model (epistemic) uncertainty~\cite{he2025survey}. A prevalent strategy is to learn probabilistic latent representations using likelihood-based generative models such as variational autoencoders (VAEs)~\cite{kingma2013auto}. Chang et al.~\cite{chang2020data} applied VAE-style modeling to jointly learn feature means and variances, enabling uncertainty estimation under noisy inputs. To model uncertainty in inter-label relationships, conditional VAEs (CVAEs)~\cite{sohn2015learning} have been adopted; Probabilistic U-Net~\cite{kohl2018probabilistic} further integrates CVAE modeling with U-Net to produce stochastic outputs.

%Recently, there are some works combining uncertainty modeling with feature representation. For example, the VAE ~\cite{kingma2013auto} , a likelihood based generative model, is trained by maximizing the evidence lower bound of the likelihood function. Specifically, some frameworks based on the VAE model have been proposed to account for the data uncertainty arising from input noise. Chang et al. ~\cite{chang2020data} applies VAE to face recognition by jointly learning feature representations (means) and their corresponding uncertainties (variances) for the first time. To capture uncertainty in inter-label relationships, the conditional VAE (CVAE) ~\cite{sohn2015learning}  framework is adopted. Probabilistic U-Net model ~\cite{kohl2018probabilistic} integrates the architectures of CVAE and U-Net, where the U-Net serves as the encoder to generate a probabilistic segmentation map.

Beyond representation learning, recent studies indicate that conventional cross-entropy training can produce overconfident predictions in classification ~\cite{wang2021can}. Motivated by evidential theory, evidential neural networks (ENNs)~\cite{sensoy2018evidential} have been explored to represent uncertainty in the output space. For example, Zhao et al.~\cite{zhao2023open} proposed an ENN that estimates multi-action uncertainty using Beta distributions. Related works also incorporate uncertainty into multi-view learning and robust cross-modal retrieval~\cite{qin2022deep}. ETMC~\cite{han2022trusted} provided a new multi-view classification paradigm with reliable integration and decision explainability. However, most existing evidential formulations are developed for single-label or multi-view classification, and are not directly aligned with the multi-label, highly imbalanced binary setting of AU detection.

%Existing studies indicate that the direct use of conventional cross-entropy loss in AU classification can lead to overconfidence in the predictions. Inspired by the success of evidential theory in the classification task, several works have explored the incorporation of uncertainty into AU classification and other related work. Zhao et al. ~\cite{zhao2023open} develop a  Evidential Neural Network (ENN) ~\cite{sensoy2018evidential} that estimates multi-action uncertainty using Beta distributions, based on actor–context–object relational representations. Bao et al. ~\cite{bao2021evidential} jointly formulate the multi-view classification and uncertainty modeling. Qin et al. ~\cite{qin2022deep} and Huang et al. ~\cite{huang2024crest} enhance cross-modal retrieval robustness through uncertainty-aware modeling of cross-modal correspondences. However, most of these methods mainly target multi-view single-label classification, which limits their applicability to multi-label binary AU detection. 

\section{Methods}

\subsection{Overview}
Given a sequence of $T$ consecutive frames, our goal is to predict the occurrence of each target AU for the central frame, together with a reliable estimate of predictive uncertainty.
As shown in Figure~\ref{fig:framework}, our UAU-Net comprises four components: (i) a spatio-temporal feature extraction module, (ii) a CVAE-based AU-specific feature extractor (CV-AFE), (iii) a global AU relationship modeling module, and (iv) an Asymmetric Beta Evidential Neural Network (AB-ENN) for uncertainty-aware AU prediction. Following common practice~\cite{wang2024multi}, we adopt a ResNet backbone to extract frame-level features.

The spatio-temporal feature extraction module aggregates facial dynamics and global spatial context across neighboring frames to form a compact representation for the target frame. Conditioned on this representation, CV-AFE maps AU-specific features to probabilistic embeddings by estimating both feature means and variances, thereby quantifying representation uncertainty. We then perform global relationship modeling to capture cross-regional inter-AU dependencies and refine AU features with contextual evidence. Finally, AB-ENN formulates AU prediction as a binary evidential learning problem with Beta distributions, producing calibrated occurrence estimates together with explicit predictive uncertainty.

\subsection{Spatio-Temporal Feature Extraction}

Facial muscle movements evolve continuously, making short-term changes between adjacent frames particularly informative for AU detection~\cite{li2021integrating}. In addition, AUs exhibit strong spatial dependencies, which should be taken into account during representation learning. To jointly capture temporal dynamics and long-range spatial context, we design a spatio-temporal feature extraction module that integrates multi-scale temporal differencing, adaptive scale-aware weighting to encode facial motion across neighboring frames, and global directional attention to model interactions across facial regions.

\paragraph{\textbf{Multi-scale Temporal Differencing and Adaptive Weighting.}} 

We first capture multi-scale facial dynamics by computing pixel-wise temporal differences between consecutive frames:
\begin{equation}
d_l^t = x_l^t - x_l^{t-1},
\label{differ}
\end{equation}
where $x_l^t$ denotes the feature map from the $l$-th backbone layer at frame $t$. 
This operation captures short-term motion patterns at different resolutions.

To enable effective cross-scale fusion, scale-specific convolution layers with kernel size and stride proportional to $8/l$ are applied to align spatial resolutions across different backbone layers. 
Temporally aggregated features within a local temporal window centered at the target frame produce summarized dynamic representations $\bar{d}_l^t$.

Rather than treating all scales equally, an adaptive weighting mechanism emphasizes AU-relevant motion cues. Temporally aggregated features from all scales are concatenated and passed through a $1\times1$ convolution followed by Softmax normalization to generate scale-wise spatial weighting maps. 
The final motion-aware feature is obtained by a weighted summation:

\begin{equation}
\mathbf{x}_{\mathrm{motion}}^t = \sum_{l=1}^{L} \mathbf{w}_l^t  \otimes \bar{d}_l^t,
\label{motion}
\end{equation}
where $\mathbf{w}_l^t$ denotes the learned scale-wise weights and $\otimes$ indicates element-wise multiplication.

\paragraph{\textbf{Global Depthwise Dual-Directional Attention (GDA)}} 
To capture long-range dependencies among facial regions, we propose a Global Depthwise Dual-Directional Attention (GDA) module. Unlike standard global pooling, GDA employs depthwise convolutions to encode global context along both horizontal and vertical directions, enabling more expressive modeling of spatial interactions that are informative for AU relationship reasoning.

Given a backbone feature map, horizontal and vertical depthwise convolutions extract directional context features. These features are concatenated and projected through a pointwise convolution to generate directional attention maps. After an element-wise combination of horizontal and vertical attention, the resulting attention is then applied to the original feature map to enhance AU-relevant regions.

Finally, we combine the motion-aware feature $\mathbf{x}_{\mathrm{motion}}^t$ with the globally attended spatial feature $\mathbf{F}'$ via element-wise summation to obtain the final spatio-temporal representation:
\begin{equation}
\mathbf{F}_t = \mathbf{x}_{\mathrm{motion}}^t + \mathbf{F}'.
\label{feature}
\end{equation}
This representation jointly encodes multi-scale facial dynamics and long-range spatial context, providing a strong foundation for uncertainty-aware AU representation learning and evidential prediction.

\subsection{CVAE-based AU-specific Feature Extractor with Uncertainty Quantification}

As discussed in \autoref{sec:intro}, uncertainty in AU detection task arises from both visual evidence (e.g., blur and occlusion) and context-dependent dependencies among AUs. We refer to these two sources as \emph{data uncertainty} and \emph{relational uncertainty}, respectively. To explicitly characterize both, we propose the CV-AFE module, which learns a probabilistic AU-specific representations by mapping deterministic features to a conditional latent distribution.

%To address these challenges, we propose the CV-AFE module (CVAE-based AU-specific Feature Extractor), which effectively captures both types of uncertainty. 

The VAE formulation captures data uncertainty by modeling the variability in the input facial features ~\cite{chang2020data}. Moreover, conditioning the latent distribution on AU labels enables CV-AFE to account for relational uncertainty induced by inter-AU dependencies, improving robustness when contextual evidence is weak or ambiguous.

%In our CV-AFE module, the extracted global feature map $\mathbf{F}_t \in \mathbb{R}^{C \times H \times W}$ is first divided into three regional subsets according to the empirically optimal configuration, each corresponding to facial areas containing different numbers of AUs: 
%(1) the upper region covering the eyebrows and eyes, 
%(2) the middle region containing the nose and cheeks, and 
%(3) the lower region encompassing the mouth and chin. 
%The specific spatial ranges of each region are defined as follows:  

Given the spatio-temporal feature map $\mathbf{F}_t \in \mathbb{R}^{C \times H \times W}$, we first partition it into three overlapping facial regions following common practice: (1) upper (brows/eyes), (2) middle (nose/cheeks), and (3) lower (mouth/chin). These three regional subsets of $\mathbf{F}_t$ are defined along the height dimension $H$ as:
\begin{equation}
\begin{aligned}
\mathbf{F}_t^\mathrm{up} &= \mathbf{F}_t[0 : \tfrac{3}{7}H], \\
\mathbf{F}_t^\mathrm{mid} &= \mathbf{F}_t[\tfrac{2}{7}H : \tfrac{5}{7}H], \\
\mathbf{F}_t^\mathrm{low} &= \mathbf{F}_t[\tfrac{4}{7}H : H],
\end{aligned}
\label{aligned}
\end{equation}

We then apply $N$ AU-specific feature extractors (AFEs), each implemented as a convolution layer followed by global average pooling (GAP). For the $n$-th AU, the extractor produces an AU-specific feature $\mathbf{x}_n^t \in \mathbb{R}^{1\times b}$ from its corresponding regional subset, where $b$ denotes the AU feature dimension. Because each $\mathbf{x}_n^t$ is computed from a region shared by multiple spatially adjacent AUs, the resulting features naturally encode intra-regional contextual cues. Consequently, each AU feature is extracted under the context of its neighboring AUs, which provides an implicit and lightweight way to model intra-regional AU dependencies at the representation stage.

Next, each AU-specific feature $\mathbf{x}_n^t$ is fed into the CVAE, allowing the model to capture both data uncertainty and relational uncertainty among AUs. Within the CVAE framework, each AU-specific feature $\mathbf{x}_n^t$ is mapped to a conditional latent distribution rather than a deterministic point in the embedding space. Specifically, the encoder network infers a Gaussian posterior over the latent variable conditioned on the input:

\begin{equation}
q_{\phi}(\mathbf{z}_n^t \mid \mathbf{x}_n^t) = \mathcal{N}(\mathbf{z}_n^t \mid \boldsymbol{\mu}_n^t, (\boldsymbol{\sigma}_n^t)^2 \mathbf{I}),
\label{distribution}
\end{equation}
where the mean $\boldsymbol{\mu}_n^t$ and standard deviation $\boldsymbol{\sigma}_n^t$ are predicted by two parallel branches of the encoder:

\begin{equation}
\boldsymbol{\mu}_n^t = f_{\theta_1}(\mathbf{x}_n^t), \quad
\boldsymbol{\sigma}_n^t = f_{\theta_2}(\mathbf{x}_n^t).
\label{meanandvariance}
\end{equation}

The posterior distribution is modeled as a diagonal multivariate Gaussian. The mean vector encodes the AU-specific representation of the $n$-th AU at frame $t$, 
while the variance explicitly quantifies the uncertainty associated with the inferred latent code. As a result, the latent representation is no longer a point estimate; instead, it is described by a stochastic latent variable drawn from the learned posterior distribution:

\begin{equation}
\mathbf{z}_n^t \sim q_{\phi}(\mathbf{z}_n^t \mid \mathbf{x}_n^t).
\label{latent}
\end{equation}

This stochastic modeling enables the network to capture intra-class variability as well as noise inherent in facial images.
Since direct sampling from the posterior $q_{\phi}(\mathbf{z}_n^t \mid \mathbf{x}_n^t)$ is not differentiable, we adopt the standard reparameterization trick~\cite{kingma2013auto}:

\begin{equation}
\mathbf{z}_n^t = \boldsymbol{\mu}_n^t + \boldsymbol{\sigma}_n^t  \boldsymbol{\epsilon}, \quad 
\boldsymbol{\epsilon} \sim \mathcal{N}(\mathbf{0}, \mathbf{I}),
\label{latent2}
\end{equation}
where $\boldsymbol{\epsilon}$ is independent noise, enabling gradient backpropagation through sampling. Given the sampled latent representation $\mathbf{z}_n^t$, an auxiliary classification head is used to supervise CV-AFE.
The classification loss is defined as:

\begin{equation}
\mathcal{L}_\mathrm{cls} = - \frac{1}{n} \sum_{i=1}^{n} \log \frac{\exp(\mathbf{w}_{y_i}^\top \mathbf{z}_i^t)}{\sum_{c=1}^{C} \exp(\mathbf{w}_c^\top \mathbf{z}_i^t)},
\label{cvae}
\end{equation}
where $\mathbf{w}_{y_i}$ denotes the classifier weight corresponding to the ground-truth label.

To prevent trivial collapse of the predicted uncertainty $\boldsymbol{\sigma}_n^t$, we regularize the latent distribution toward a standard normal prior using the KL divergence:

\begin{equation}
\mathcal{L}_\mathrm{KL} = \mathrm{KL}\big(\mathcal{N}(\boldsymbol{\mu}_n^t, (\boldsymbol{\sigma}_n^t)^2) \,\|\, \mathcal{N}(\mathbf{0}, \mathbf{I})\big).
\end{equation}

The total training loss combines the classification and KL losses:

\begin{equation}
\mathcal{L}_\mathrm{CV-AFE} = \mathcal{L}_\mathrm{cls} + \lambda_1 \mathcal{L}_\mathrm{KL}.
\label{cvaeloss}
\end{equation}
where $\lambda_1$ is a trade-off hyperparameter balancing the classification loss and the KL divergence.

In addition to feature-level modeling, spatio-temporal relationships
among AUs within the same facial region are further captured at the
intra-regional AU dependency modeling. To this end, an ACP (AU combination prediction) is introduced during training.
For each sliced regional feature
$\mathbf{F}_t^{\text{sub}} \in \{\mathbf{F}_t^{\text{up}}, 
\mathbf{F}_t^{\text{mid}}, \mathbf{F}_t^{\text{low}}\}$, the ACP branch
predicts a joint AU occurrence pattern
$\mathbf{Y}_t^{\text{sub}} = \{y_{t,1}^{\text{sub}}, \dots,
y_{t,2^{N_{\text{sub}}}}^{\text{sub}}\}$, where $N_{\text{sub}}$ denotes the
number of target AUs in the corresponding facial sub-region.

Formally, the AU combination prediction is defined as
\begin{equation}
\mathbf{P}_t^{\text{sub}} =
\mathrm{Softmax}\big(
\mathrm{FC}_{\text{sub}}(\mathrm{GAP}(\mathbf{F}_t^{\text{sub}}))
\big),
\end{equation}
where $\mathrm{FC}_{\text{sub}}(\cdot)$ denotes a fully connected layer.
Training this branch encourages the network to encode local AU
co-occurrence patterns into regional features, thereby enhancing
AU-relevant representations.

After the first-stage training, our CV-AFE module extracts the feature $\mathbf{v}_n^t$ for each AU ($k_1$ AUs for the up region), while simultaneously capturing both data uncertainty and relational uncertainty among AUs, enhancing the model's robustness to noise and the relationship uncertainty.

\subsection{Global Relationship Modeling}

Beyond intra-regional dependencies, AU activations can correlate across facial regions~\cite{tong2007facial}. To capture such cross-regional relationships, we construct a graph at each time step $t$ in which each AU-specific feature $\mathbf{v}_n^t$ is treated as a node. We then adaptively connect each node to a subset of AUs from other region that are predicted to be activated by the ACP branch. This design is motivated by the observation that activated AUs provide stronger and more reliable contextual cues for reasoning about other AUs, while filtering out spurious connections induced by irrelevant or inactive AUs.

The cross-regional relationships are modeled using a Graph Attention Network (GAT)~\cite{2017Graph}. The attention coefficient between the $n$-th and $m$-th AU nodes at time $t$ is computed as:

\begin{equation}
e_{n,m}^t =
\mathrm{LeakyReLU}
\big(
\mathbf{r}^\top
[ \mathbf{W}\mathbf{v}_n^t \,\|\, \mathbf{W}\mathbf{v}_m^t ]
\big),
\end{equation}
followed by normalized attention weights:

\begin{equation}
\alpha_{n,m}^t =
\frac{\exp(e_{n,m}^t)}
{\sum_{q \in \mathcal{N}_n^t} \exp(e_{n,q}^t)},
\end{equation}
where $\mathbf{W} \in \mathbb{R}^{b \times b}$ is a shared linear
transformation, $\mathbf{r} \in \mathbb{R}^{2b}$ denotes the attention
weight vector, $\|\,$ represents concatenation, and
$\mathcal{N}_n^t$ denotes the neighbor set of node $n$.

The updated AU representation is obtained via:

\begin{equation}
\hat{\mathbf{v}}_n^t =
\phi\!\left(
\sum_{m \in \mathcal{N}_n^t}
\alpha_{n,m}^t \mathbf{W}\mathbf{v}_m^t
\right),
\end{equation}
where $\phi(\cdot)$ denotes a non-linear activation function.

The GAT outputs are then passed to a temporal convolutional network (TCN)
to aggregate temporal context and obtain a set of spatio-temporal
relationship-aware AU features $\hat{\mathbf{V}}_t = \{\hat{\mathbf{v}}_1^t, \dots, \hat{\mathbf{v}}_N^t\}$ for frame $t$.

%Consequently, After being processed by the TCN, a set of hierarchical relationship-aware AU features $\hat{\mathbf{V}}_t = \{\hat{\mathbf{v}}_1^t, \dots, \hat{\mathbf{v}}_n^t\}$ is obtained to represent all target AUs in frame $t$.

\subsection{Asymmetric Beta-ENN for Multi-label AU Classification}

In this section, multi-label AU classification is performed after obtaining the relationship-aware feature for each AU in frame $t$. To explicitly quantify predictive uncertainty and mitigate overconfident point estimates of conventional methods, we adopt an evidential learning formulation.

This choice is motivated by the observation that the predicted likelihood of AU occurrence is a Bernoulli (binomial) variable, whose conjugate prior is a Beta distribution. Accordingly, we design an \emph{Asymmetric Beta Evidential Neural Network} (AB-ENN) that models each AU as a binary evidential learning problem with a Beta posterior.
  
Specifically, for the $n$-th AU, there are two mutually exclusive hypotheses (active/inactive) with associated masses $b_n, d_n, u_n$, where $b_n$ and $d_n$ are the belief masses for positive and negative labels, and $u_n$ denotes the uncertainty mass. These masses satisfy $b_n + d_n + u_n = 1$ and $b_n, d_n, u_n \ge 0$.

Given a prediction function $f(\cdot)$, for an instance $\mathbf{x}$, we obtain evidences $(e_n^+, e_n^-) = f_n(\mathbf{x})$ for the positive and negative classes. Based on these evidences, the corresponding masses are given as:

\begin{equation}
b_n = \frac{e_n^+}{S_n}, \quad
d_n = \frac{e_n^-}{S_n}, \quad
u_n = \frac{2}{S_n}
\end{equation}
where $S_n = e_n^+ + e_n^- + 2$ refers to the Beta strength.

The predicted probability $p_n \in [0,1]$ for the $n$-th AU is modeled by a Beta distribution, $\mathrm{Beta}(p_n \mid \alpha_n, \beta_n)$, with parameters $\alpha_n = e_n^+ + 1$ and $\beta_n = e_n^- + 1$.  
The expected probability for the $n$-th AU is then the mean of the Beta distribution:
$\bar{p}_n =  \frac{\alpha_n}{\alpha_n + \beta_n}.$

Although evidential binary cross-entropy (EBCE)~\cite{sensoy2018evidential} has achieved promising results in evidential learning, its performance in AU detection is limited by the severe class imbalance of AU datasets, where negative labels dominate and positive labels are sparse. To mitigate this, we adopt the Asymmetric Focal Loss (ASL) ~\cite{9710171}, which down-weights easy negatives and emphasizes positive classes. For a predicted probability $p$ and label $y \in \{0,1\}$, ASL is defined as:

%\begin{equation}
%\mathcal{L}_{\text{ASL}}(p, y) =
%\begin{cases}
%-(1-p)^{\gamma^+} \log(p), & y = 1,\\
%-(p-m)^{\gamma^-} \log(1-p), & y = 0,
%\end{cases}
%\label{asl}
%\end{equation}
\begin{equation}
\mathcal{L}_{\text{ASL}}(p, y) =
\left\{
\begin{aligned}
&-(1-p)^{\gamma^+} \log(p), && y = 1, \\
&-(p-m)^{\gamma^-} \log(1-p), && y = 0.
\end{aligned}
\right.
\label{asl}
\end{equation}
where $m = \min(p, c)$ is a small shift to ignore very easy negatives, $c$ is a constant probability shift parameter used for neglecting very easy negative examples, and $\gamma^+, \gamma^-$ are focus parameters, typically set with $\gamma^- > \gamma^+$ to emphasize positive examples.

Following the same evidential principle, we define the Asymmetric Beta Loss (ABL) by taking the Bayesian risk of ASL under the Beta-distributed prediction. For $y = 1$, the integral admits a closed-form solution. For $y = 0$, the probability-shifting technique requires a modification since the output is a distribution rather than a point estimate.  
For the $n$-th AU, we define a shifted random variable $p_{nc} \sim \mathrm{Beta}(\alpha_{nc}, \beta_n)$ with $\alpha_{nc} = \max\left( \alpha_n - \frac{c}{1-c} \, \beta_n, \; 0 \right)$. 

The ABL for frame $t$ is defined as follows:
%{\small
%\begin{equation}
%\mathcal{L}_{\text{AB}}^t =
%\begin{cases}
%- w_n\displaystyle \int_0^1 (1 - p_n^t)^{\gamma^+} \log(p_n^t)\, \mathrm{Beta}(p_n^t \mid \alpha_n^t, \beta_n^t)\, dp_n^t, & y_n^t = 1,\\
%- w_n\displaystyle \int_0^1 (p_{nc}^t)^{\gamma^-} \log(1 - p_{nc}^t)\, \mathrm{Beta}(p_{nc}^t \mid \alpha_{nc}^t, \beta_n^t)\, dp_{nc}^t, & y_n^t = 0,
%\end{cases}
%\end{equation}
%}
{\footnotesize
\begin{equation}
\mathcal{L}_{\text{AB}}^t =
\left\{
\begin{aligned}
&- w_n \int_0^1 (1 - p_n^t)^{\gamma^+} \log(p_n^t)\,
\mathrm{Beta}(p_n^t \mid \alpha_n^t, \beta_n^t)\, dp_n^t,& y_n^t = 1, \\
&- w_n \int_0^1 (p_{nc}^t)^{\gamma^-} \log(1 - p_{nc}^t)\,
\mathrm{Beta}(p_{nc}^t \mid \alpha_{nc}^t, \beta_{n}^t)\, dp_{nc}^t,& y_n^t = 0.
\end{aligned}
\right.
\end{equation}
}
which admits the analytical solution:

%\begin{equation}
%\mathcal{L}_{\text{AB}}^t =
%\begin{cases}
%w_nw^+ \big[ \psi(\alpha^t_n + \beta^t_n + \gamma^+) - \psi(\alpha_n^t) \big], & y_n^t = 1, \\[1ex]
%w_nw^- \big[ \psi(\alpha_{nc}^t + \beta^t_n + \gamma^-) - \psi(\beta_n^t) \big], & y_n^t = 0,
%\end{cases}
%\end{equation}

\begin{equation}
\mathcal{L}_{\text{AB}}^t =
\left\{
\begin{aligned}
&w_n w^+ \big[ \psi(\alpha_n^t + \beta_n^t + \gamma^+) - \psi(\alpha_n^t) \big], && y_n^t = 1, \\
&w_n w^- \big[ \psi(\alpha_{nc}^t + \beta_n^t + \gamma^-) - \psi(\beta_n^t) \big], && y_n^t = 0.
\end{aligned}
\right.
\end{equation}
where $w_n$ is calculated for each AU based on the training set, the weighting coefficients $w^+$ and $w^-$ are formulated as:

\begin{equation}
\begin{aligned}
w^+ &= \sum_{r=0}^{\gamma^+ - 1} \frac{\beta^t_n + r}{\alpha_n^t + \beta_n^t + r}, \\
w^- &= \sum_{r=0}^{\gamma^- - 1} \frac{\alpha^t_n + r}{\alpha_{nc}^t + \beta_n^t + r}.
\end{aligned}
\end{equation}

To reduce the influence of highly uncertain predictions, we down-weight the per-frame loss using the predicted uncertainty mass, thereby emphasizing more reliable samples. Specifically, for frame $t$, we compute the uncertainty as $u^t = \frac{2N}{\sum_{j=1}^{N} (\alpha_j^t + \beta_j^t)}$ and weight the loss by $(1-u^t)$.

Finally, the overall ABL ${L}_{\text{AB}}$ is defined as:
\begin{equation}
\begin{aligned}
%\mathcal{L}_{\text{AB}} = \sum_{t=1}^{T}(1-u^t)\mathcal{L}^t_{\text{AB}}
\mathcal{L}_{\text{AB}} = \sum_{t=1}^{T} (1-u^t)\,\mathcal{L}^t_{\text{AB}}
\end{aligned}
\end{equation}

In addition, we apply a cross-entropy loss ${L}_{\text{sub}}$ to supervise the regional AU combination predictions for each frame in the first stage of the ACP module:

\begin{equation}
\mathcal{L}_{\text{sub}} = \sum_{t=1}^{T} \sum_{\text{sub} \in \{\text{up}, \text{mid}, \text{low}\}} 
\text{CE}(P_{\text{sub}}^t, Y_{\text{sub}}^t),
\end{equation}
where $P_{\text{sub}}^t$ and $Y_{\text{sub}}^t$ denote the predicted AU combination and the corresponding ground-truth for a given facial region in frame $f_t$, and $\text{CE}(\cdot)$ is the cross-entropy function. 

By predicting combinations of multiple AUs within the same facial region, the network is encouraged to capture local co-occurrence patterns among spatially adjacent AUs. Consequently, the overall loss function for the second training stage combines the two losses as:
\begin{equation}
\mathcal{L} = \mathcal{L}_{\text{AB}} + \lambda_2 \, \mathcal{L}_{\text{sub}},
\label{lossbeta}
\end{equation}
where $\lambda_2$ balances the contributions of the two losses.

\section{Experiments}
\subsection{Experimental Setup}

\subsubsection{Datasets}

We evaluate the proposed method on two widely used facial action unit benchmarks, BP4D~\cite{zhang2014bp4d} and DISFA~\cite{mavadati2013disfa}. Both datasets provide frame-level AU occurrence annotations and are widely used to benchmark AU detection methods, enabling a fair and comprehensive evaluation.

\begin{itemize}
\item \textbf{BP4D}~\cite{zhang2014bp4d} consists of 328 facial video sequences collected from 41 subjects (23 females and 18 males), resulting in approximately 140,000 annotated frames. Each frame is labeled with the occurrence status of 12 AUs, covering a diverse range of spontaneous facial expressions.

\item \textbf{DISFA}~\cite{mavadati2013disfa} contains 27 facial video sequences with a total of 130,815 frames, recorded from 27 subjects (12 females and 15 males) while they watched online videos. Compared with BP4D, DISFA focuses on 8 AUs and exhibits more subtle facial motions, making AU detection particularly challenging.

\end{itemize}
\subsubsection{Implementation Details}

Following prior approaches in AU detection~\cite{zhang2018classifier}, we employ MTCNN~\cite{yang2023fan} to detect, crop, and align faces in each frame. Specifically, a face area of size $224 \times 224$ is extracted for subsequent processing. For both datasets, we adopt subject-independent three-fold cross-validation, and the final performance is reported by averaging the results over all three folds.

To ensure temporal consistency, we pad each input sequence by duplicating the first and last frames at the beginning and end of each video, respectively, so that all frames can be processed uniformly by the proposed model. The network is optimized using the AdamW~\cite{loshchilovdecoupled} optimizer with parameters $\beta_1 = 0.9$ and $\beta_2 = 0.999$.

The training procedure consists of two stages. In Stage~I, we train CV-AFE independently to learn uncertainty-aware AU representations, where the weighting parameter $\lambda_1$ in Eq.~\ref{cvaeloss} is set to $0.01$. In Stage~II, we jointly optimize the local and global AU classifiers; we set the trade-off parameter $\lambda_2$ in Eq.~\ref{lossbeta} to $0.01$, and the probability shift parameter $c$ in Eq.~\ref{asl} to $0.2$. A cosine decay learning-rate scheduler is adopted with an initial learning rate of $1 \times 10^{-4}$. The backbone network is initialized with ImageNet pretraining~\cite{deng2009imagenet}.

All experiments are conducted on NVIDIA RTX 4090 GPUs.

\subsubsection{Evaluation Metric}

Consistent with prior AU recognition studies, frame-level F1-score is adopted as the primary evaluation metric, defined as
\begin{equation}
\mathrm{F1} = \frac{2PR}{P + R},
\end{equation}
where $P$ and $R$ denote precision and recall, respectively. Under class imbalance, this metric captures the trade-off between performance on both positive (active) and negative (inactive) predictions.

\begin{table*}[t]
\small 
\centering
\caption{Frame-based F1-score (\%) comparison on the BP4D dataset.
The best result for each AU and the overall average are highlighted in \textbf{bold}.}
\label{tab:bp4d_comparison}
%\begin{tabularx}{\textwidth}{l *{12}{>{\centering\arraybackslash}X} >{\centering\arraybackslash}X}
\begin{tabular*}{\textwidth}{@{\extracolsep{\fill}} lccccccccccccc}
\toprule
Method & AU1 & AU2 & AU4 & AU6 & AU7 & AU10 & AU12 & AU14 & AU15 & AU17 & AU23 & AU24 & Avg \\
\midrule
EAC-Net ~\cite{li2018eac} & 39.0 & 35.2 & 48.6 & 76.1 & 72.9 & 81.9 & 86.2 & 58.8 & 37.5 & 59.1 & 35.9 & 35.8 & 55.6 \\
JAA-Net ~\cite{shao2021jaa} & 47.2 & 44.0 & 54.9 & 77.5 & 74.6 & 84.0 & 86.9 & 61.9 & 43.6 & 60.3 & 42.7 & 41.9 & 60.0 \\
ARL ~\cite{shao2019facial} & 45.8 & 39.8 & 55.1 & 75.7 & 77.2 & 82.3 & 86.6 & 58.8 & 47.6 & 62.1 & 47.4 & 55.4 & 61.2 \\
SMA-Net ~\cite{li2021your} & 56.5 & 45.1 & 57.0 & 79.5 & 79.5 & 84.5 & 86.4 & 66.1 & 55.8 & 64.2 & 48.7 & 56.8 & 65.0 \\
SRERL ~\cite{li2019semantic} & 46.9 & 45.3 & 55.6 & 77.1 & 78.4 & 83.5 & 87.6 & 63.9 & 52.2 & 63.9 & 47.1 & 53.3 & 62.9 \\
%FAUDT ~\cite{jacob2021facial} & 51.7 & 49.3 & 61.0 & 77.8 & 79.5 & 82.9 & 86.3 & 67.6 & 51.9 & 63.0 & 43.7 & 56.3 & 64.2 \\
ME-GraphAU ~\cite{ijcai2022p173} & 52.7 & 44.3 & 60.9 & 79.9 & 80.1 & 85.3 & 89.2 & 69.4 & 55.4 & 64.4 & 49.8 & 55.1 & 65.5 \\
SACL ~\cite{liu2024multi} & 57.8 & 48.8 & 59.4 & 79.1 & 78.8 & 84.0 & 88.2 & 65.2 & 56.1 & 63.8 & 50.8 & 55.2 & 65.6 \\
AC2D ~\cite{shao2025facial} & 54.2 & \textbf{54.7} & 56.5 & 77.0 & 76.2 & 84.0 & 89.0 & 63.6 & 54.8 & 63.6 & 46.5 & 54.8 & 64.6 \\
AUNCE ~\cite{SHANG2026112746} & 53.6 & 49.8 & 61.6 & 78.4 & 78.8 & 84.7 & 89.6 & 67.4 & 55.1 & 65.4 & 50.9 & 58.0 & 66.1 \\
%STRAL ~\cite{shao2020spatio} & 48.2 & 47.7 & 58.1 & 75.8 & 78.1 & 81.6 & 87.6 & 60.5 & 50.2 & 64.0 & 51.2 & 55.2 & 63.2 \\
HSTR-Net ~\cite{song2022heterogeneous} & 55.5 & 49.5 & \textbf{61.9} & 76.6 & 80.2 & 84.2 & 87.4 & 62.6 & 54.8 & 64.1 & 47.1 & 52.1 & 64.7 \\
WSRTL ~\cite{yan2022weakly} & \textbf{59.7} & 51.7 & 61.6 & \textbf{80.3} & \textbf{80.9} & 85.2 & \textbf{89.7} & 67.8 & 52.2 & 63.4 & \textbf{51.4} & 46.9 & 65.9 \\
MDHR ~\cite{wang2024multi} & 58.3 & 50.9 & 58.9 & 78.4 & 80.3 & 84.9 & 88.2 & \textbf{69.5} & 56.0 & \textbf{65.5} & 49.5 & 59.3 & 66.6 \\
UGN ~\cite{song2021uncertain} & 54.2 & 46.4 & 56.8 & 76.2 & 76.7 & 82.4 & 86.1 & 64.7 & 51.2 & 63.1 & 48.5 & 53.6 & 63.3\\
AU-TTT ~\cite{11209184} & -- & -- & -- & -- & -- & -- & -- & -- & -- & -- & -- & -- & 65.6 \\
RETD ~\cite{chang2024facial} & 54.7 & 50.8 & 57.1 & 78.8 & 79.6 & 84.6 & 88.0 & 67.0 & 54.9 & 62.9 & 48.6 & 54.5 & 65.1 \\
\midrule
Ours & 57.9 & 51.3 & \textbf{61.9} & 79.9 & 78.9 & \textbf{85.9} & 88.8 & 65.9 & \textbf{56.3} & 64.4 & 51.3 & \textbf{60.1} & \textbf{66.8} \\
\bottomrule
\end{tabular*}
\end{table*}

\begin{table*}[t]
\small
\centering
\caption{Frame-based F1-score (\%) comparison on the DISFA dataset.
Best results for each AU are highlighted in \textbf{bold}.}
\label{tab:disfa_comparison}

%\begin{tabularx}{\textwidth}{l *{8}{>{\centering\arraybackslash}X} >%{\centering\arraybackslash}X}
\begin{tabular*}{\textwidth}{@{\extracolsep{\fill}} lccccccccc}
\toprule
Method 
& AU1 & AU2 & AU4 & AU6 & AU9 & AU12 & AU25 & AU26 & Avg \\
\midrule

EAC-Net ~\cite{li2018eac}
& 41.5 & 26.4 & 66.4 & 50.7 & \textbf{80.5} & \textbf{89.3} & 88.9 & 15.6 & 48.5 \\

JAA-Net ~\cite{shao2021jaa}
& 43.7 & 46.2 & 56.0 & 41.4 & 44.7 & 69.6 & 88.3 & 58.4 & 56.0 \\

ARL ~\cite{shao2019facial}
& 43.9 & 42.1 & 63.6 & 41.8 & 40.0 & 76.2 & 95.2 & 66.8 & 58.7 \\

SMA-Net ~\cite{li2021your}& 53.4 & 54.2 & 64.0 & 57.0 & 47.0 & 76.6 & 92.0 & 55.2 & 64.2 \\

SRERL ~\cite{li2019semantic}& 45.7 & 47.8 & 59.6 & 47.1 & 45.6 & 73.5 & 84.3 & 43.6 & 55.9 \\

%FAUDT ~\cite{jacob2021facial}& 46.1 & 48.6 & 72.8 & 56.7 & 50.0 & 72.1 & 90.8 & 55.4 & 61.5 \\

ME-GraphAU ~\cite{ijcai2022p173}
& 54.6 & 47.1 & 72.9 & 54.0 & 55.7 & 76.7 & 91.1 & 53.0 & 63.1 \\

SACL ~\cite{liu2024multi}
& 62.0 & \textbf{65.7} & 74.5 & 53.2 & 43.1 & 76.9 & 95.6 & 53.1 & 65.5 \\

AC2D ~\cite{shao2025facial}
& 57.8 & 59.2 & 70.1 & 50.1 & 54.4 & 75.1 & 90.3 & 66.2 & 65.4 \\

AUNCE ~\cite{SHANG2026112746}
& 61.8 & 58.9 & 74.9 & 49.7 & 56.2 & 73.5 & 92.1 & 64.2 & 66.4 \\

%STRAL ~\cite{shao2020spatio}& 52.2 & 47.4 & 68.9 & 47.8 & 56.7 & 72.5 & 91.3 & 67.6 & 63.0 \\

HSTR-Net ~\cite{song2022heterogeneous}
& 54.3 & 50.8 & 70.1 & \textbf{66.6} & 59.6 & 68.0 & \textbf{97.9} & \textbf{69.8} & 62.9 \\

WSRTL ~\cite{yan2022weakly}
& 57.3 & 51.8 & 74.3 & 49.8 & 44.8 & 79.3 & 94.6 & 64.6 & 64.6 \\

MDHR ~\cite{wang2024multi}
& 65.4 & 60.2 & 75.2 & 50.2 & 52.4 & 74.3 & 93.7 & 58.2 & 66.2 \\

UGN ~\cite{song2021uncertain}& 43.3 & 48.1 & 63.4 & 49.5 & 48.2 & 72.9 &  90.8 &  59.0 & 60.0 \\

AU-TTT ~\cite{11209184}
& -- & -- & -- & -- & -- & -- & -- & -- & 66.4 \\

RETD ~\cite{chang2024facial}
& 62.6 & 54.7 & 70.8 & 46.3 & 51.7 & 76.3 & 94.4 & 59.8 & 64.6 \\

\midrule
\textbf{Ours}
& \textbf{66.3} & 58.4 & \textbf{75.7} & 53.4 & 52.0 & 72.6 & 91.8 & 62.5 & \textbf{66.6} \\
\bottomrule

\end{tabular*}
\end{table*}

\subsection{Comparison with State-of-the-Art Methods}
\autoref{tab:bp4d_comparison} and \autoref{tab:disfa_comparison} compare UAU-Net with representative AU detection methods under the same subject-independent three-fold cross-validation protocol. Various modeling paradigms include static image-based models (e.g., EAC-Net~\cite{li2018eac}, JAA-Net~\cite{shao2021jaa}, ARL~\cite{shao2019facial}), static AU-relation modeling methods(e.g., SRERL~\cite{li2019semantic}, ME-GraphAU~\cite{ijcai2022p173}, SACL~\cite{liu2024multi}, AC2D~\cite{shao2025facial}, AUNCE~\cite{SHANG2026112746}), spatio-temporal approaches (e.g., %STRAL~\cite{shao2020spatio}, 
HSTR-Net~\cite{song2022heterogeneous}, WSRTL~\cite{yan2022weakly}, MDHR~\cite{wang2024multi}), the uncertainty-aware graph neural network model UGN~\cite{song2021uncertain}, as well as more recent extensions such as test-time training (AU-TTT~\cite{11209184}) and text-enhanced modeling (RETD~\cite{chang2024facial}).

\paragraph{\textbf{Overall performance}}
Our UAU-Net achieves the best average frame-based F1-score on both benchmarks, reaching \textbf{66.8\%} on BP4D and \textbf{66.6\%} on DISFA. Compared with the strongest published competitors reported in the tables, our method yields consistent gains in the overall average (i.e., \textbf{+0.2} on DISFA and \textbf{+0.3} on BP4D). While these margins are modest---which is expected given that performance on these datasets is already approaching saturation---the improvement is observed on \emph{both} datasets and under the same subject-independent evaluation setting, suggesting that the proposed uncertainty modeling brings complementary benefits beyond architectural choices.

\paragraph{\textbf{Per-AU analysis and robustness.}}
In addition to improving the overall mean, UAU-Net shows competitive or best performance on multiple individual AUs across the two datasets. Notably, the gains are more pronounced on AUs that are challenging because of subtle appearance changes and/or low positive rates (e.g., AU24 on BP4D, and AU1/AU4 on DISFA), indicating that explicitly accounting for uncertainty helps reduce overconfident errors on ambiguous samples. On AUs where competing methods already perform strongly (e.g., highly frequent AUs), our method remains competitive, suggesting that uncertainty-aware training does not sacrifice performance on easier labels.

%As shown in the tables, our method achieves state-of-the-art performance on both datasets. Specifically, it obtains an average frame-based F1-score of \textbf{66.6\%} on DISFA and \textbf{66.8\%} on BP4D, outperforming all existing approaches. Compared with traditional static image-based methods, our model demonstrates a substantial performance improvement. Moreover, when compared with the strongest existing competitors, our approach further improves the overall F1-score by \textbf{0.2\%} on DISFA and \textbf{0.3\%} on BP4D, respectively.

%These results indicate that our model is able to effectively capture the uncertainty inherent in AU detection, which arises in both the feature representation and the AU classification stage. By discouraging the model from overfitting samples and AU labels with high uncertainty, more robust representations can be learned, leading to consistent performance gains. This capability constitutes the key advantage of our approach over previous methods.

\begin{table}[t]
\centering
\caption{Ablation study of different components on the DISFA dataset (frame-based F1-score, \%).}
\label{tab:ablation}
\begin{tabular}{ccccc}
\toprule
Baseline & CV-AFE & Beta-ENN & AB-ENN & F1-score (\%) \\
\midrule
\checkmark
&  &  &  & 65.2 \\

\checkmark
& \checkmark &  &  & 66.1 \\

\checkmark
&  & \checkmark &  & 65.7 \\

\checkmark
&  &  & \checkmark & 66.1 \\

\checkmark
& \checkmark & \checkmark &  & 66.4 \\

\checkmark
& \checkmark &  & \checkmark & \textbf{66.6} \\

\bottomrule
\end{tabular}
\end{table}

\subsection{Ablation Study}
We conduct comprehensive ablation experiments on DISFA~\cite{mavadati2013disfa} dataset to evaluate the effectiveness of each proposed component in UAU-Net. All variants are trained and evaluated under the same subject-independent three-fold cross-validation protocol, and we report the average frame-based F1-score in \autoref{tab:ablation}. The baseline corresponds to the same backbone and AU prediction pipeline but \emph{without} uncertainty-aware representation learning (CV-AFE) and \emph{without} evidential classification (ENN).

\paragraph{\textbf{Effect of uncertainty-aware representation learning}}
Adding CV-AFE on top of the baseline improves the average F1 from 65.2\% to 66.1\% (\textbf{+0.9}).
This gain supports our motivation that modeling feature-level uncertainty helps the network learn more stable AU representations by reducing the impact of noisy frames and ambiguous AU co-occurrences, which are prevalent in DISFA.

%By introducing the CV-AFE module, the average F1-score is improved by 0.9\%, demonstrating that explicitly modeling uncertainty during feature extraction helps the model avoid overfitting to highly uncertain samples, thereby enhancing recognition accuracy. 

\paragraph{\textbf{Effect of evidential classification}}
Replacing the conventional point-estimation classifier with Beta-ENN alone increases the F1-score to 65.7\% (\textbf{+0.5} over the baseline).
This result suggests that explicitly parameterizing a predictive distribution (Beta posterior) yields better-calibrated confidence and reduces overconfident errors, which is particularly beneficial when positive AU labels are sparse.

%Furthermore, replacing the conventional point-estimation classifier with the Beta-ENN module in the AU prediction stage yields an additional performance gain of about 0.3\% in terms of F1-score. This improvement indicates that Beta-ENN effectively mitigates the overconfidence issue during prediction by incorporating uncertainty-aware decision making, leading to more robust AU classification.

\paragraph{\textbf{Effect of Asymmetric loss for imbalance}}
Using AB-ENN further boosts performance to 66.1\% (\textbf{+0.9} over the baseline), matching the gain brought by CV-AFE when applied individually.
Compared with Beta-ENN, AB-ENN brings an additional \textbf{+0.4} improvement (66.1 vs.\ 65.7), indicating that the asymmetric treatment of positives and negatives better aligns the training signal with the long-tailed AU label distribution.

%To further address the severe class imbalance commonly observed in AU detection datasets, the proposed AB-ENN module brings a further improvement of 0.2\% in F1-score. This verifies that the asymmetric loss design assigns differentiated feedback to positive and negative samples, which alleviates the imbalance of AU label distributions to a certain extent.

\paragraph{\textbf{Complementarity of the modules}}
Combining CV-AFE with Beta-ENN yields 66.4\%, and combining CV-AFE with AB-ENN achieves the best result of \textbf{66.6\%}. Overall, integrating both uncertainty-aware representation learning and evidential classification leads to a \textbf{+1.4} absolute gain over the baseline (66.6\% vs.\ 65.2\%).
The consistent improvements across combinations indicate that CV-AFE and (AB-)ENN address different sources of uncertainty (representation vs.\ decision), and are therefore complementary rather than redundant.

%Overall, integrating all proposed components results in an absolute F1-score improvement of approximately 1.3\%. Importantly, no negative interference is observed among the modules, indicating that they are complementary and jointly contribute to the performance improvement of the proposed model.

\subsection{Analysis of Uncertainty Impact}
This section analyzes how the proposed uncertainty modeling influences training behavior and final AU detection performance on DISFA.
We leverage the feature uncertainty estimated by CV-AFE (i.e., the learned posterior variance) to stratify training frames into three groups with \emph{low}, \emph{medium}, and \emph{high} uncertainty~\footnote{We form the three groups by splitting samples according to the uncertainty quantiles computed on the training set.}. Intuitively, low-uncertainty samples correspond to clear facial evidence and reliable AU co-occurrence cues, while high-uncertainty samples are more likely to contain occlusions, motion blur, extreme poses, or ambiguous/weak AU activations.

\begin{figure}
    \centering
    \includegraphics[width=1\linewidth]{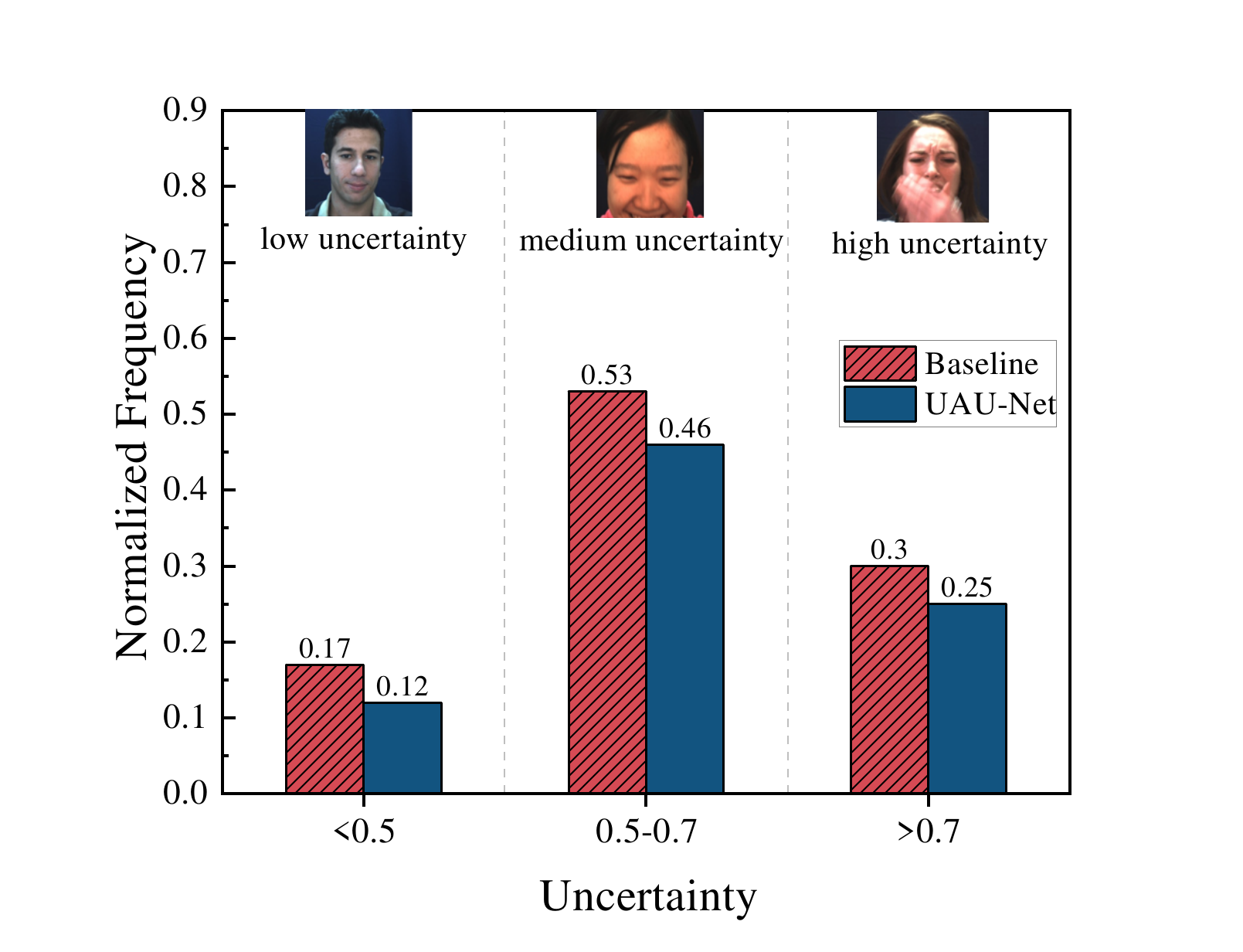}
    \caption{Failure-case analysis of the baseline and UAU-Net}
    \Description{Comparison of representative failure cases from the baseline model and UAU-Net across different uncertainty levels.}
    \label{fig:placeholder}
\end{figure}

%The motivation of this experiment is to investigate how uncertainty affects the training dynamics of the model and the final recognition performance. Based on the uncertainty estimated by the CVAE, we divide the training samples in the DISFA dataset into three categories according to their feature uncertainty: low uncertainty samples, medium uncertainty samples, and high uncertainty samples.

\paragraph{\textbf{Error distribution across uncertainty strata}}
We first compare the normalized distribution of misclassified labels produced by the baseline and by UAU-Net (see \autoref{fig:placeholder}). UAU-Net yields fewer errors on the low- and medium-uncertainty strata across AUs, indicating that explicitly modeling uncertainty improves the effective learning signal for samples that are inherently learnable. In contrast, the relative share of errors on the high-uncertainty stratum becomes larger after introducing uncertainty modeling. Importantly, this does \emph{not} imply that the model degrades on hard samples; rather, it reflects a shift in where the remaining errors concentrate once performance on easier strata improves.

%We first analyze the number and sources of misclassified labels after incorporating uncertainty modeling. As illustrated in the normalized error distribution, the qualification of uncertainty consistently improves the AU detection accuracy across all action units. Notably, compared with the baseline model, our uncertainty-aware model produces significantly fewer erroneous predictions on easy and semihard samples, indicating that explicitly modeling uncertainty enables the model to better learn discriminative representations for samples that are inherently learnable.

\paragraph{\textbf{Interpretation and training dynamics}}
This behavior is consistent with our design: both CV-AFE and AB-ENN reduce the influence of unreliable evidence (e.g., by down-weighting highly uncertain predictions in the loss), thereby preventing the network from being dominated by noisy or ambiguous frames. As a result, the model can focus more on learning robust AU patterns instead of overfitting high-uncertainty outliers, which improves overall generalization and leads to more reliable predictions.

%We further analyze the proportion of misclassified samples from each category among all misclassified instances. The results show that, compared with the baseline, our model yields a higher proportion of misclassified labels in the predictions for high-uncertainty samples. This observation implies that the proposed framework intentionally reduces its focus on highly noisy and unreliable samples instead of overfitting them. This behavior is consistent with the theoretical motivation of the framework, which aims to model uncertainty in AU feature representation and classification to prevent the network from being dominated by noisy or ambiguous training samples.

\section{Conclusion}
In this paper, we proposed UAU-Net, an uncertainty-aware framework for facial action unit detection that explicitly models uncertainty at both the representation and decision stages. At the representation stage, CV-AFE learns probabilistic AU embeddings by estimating feature means and variances, and captures relational uncertainty via label-conditioned modeling of inter-AU dependencies. At the decision stage, AB-ENN performs evidential multi-label prediction with Beta distributions and an asymmetric Beta loss to mitigate overconfident predictions under severe class imbalance. Extensive experiments on BP4D and DISFA demonstrate that UAU-Net achieves state-of-the-art performance under the standard subject independent protocol. Ablation studies verify the effectiveness and complementarity of CV-AFE and (AB-)ENN, and the uncertainty impact analysis further shows that our framework improves learning robustness by concentrating errors on genuinely high-uncertainty samples rather than overfitting them. Overall, UAU-Net offers a practical uncertainty-aware perspective for building robust AU detectors and provides a useful reference point for future research on reliable facial behavior understanding.


\bibliography{reference}

@inproceedings{ijcnnCaoLZ22,
  author       = {Jiyuan Cao and
                  Zhilei Liu and
                  Yong Zhang},
  title        = {Cross-subject Action Unit Detection with Meta Learning and Transformer-based
                  Relation Modeling},
  booktitle    = {International Joint Conference on Neural Networks, {IJCNN} 2022, Padua,
                  Italy, 2022},
  pages        = {1--8},
  publisher    = {{IEEE}},
  year         = {2022},
}

@inproceedings{icasspSongL23,
  author       = {Juan Song and
                  Zhilei Liu},
  title        = {Self-Supervised Facial Action Unit Detection with Region and Relation
                  Learning},
  booktitle    = {{IEEE} International Conference on Acoustics, Speech and Signal Processing
                  {ICASSP} 2023, Rhodes Island, Greece, June 4-10, 2023},
  pages        = {1--5},
  publisher    = {{IEEE}},
  year         = {2023},
 }

@article{li2022deep,
  title={Deep learning for micro-expression recognition: A survey},
  author={Li, Yante and Wei, Jinsheng and Liu, Yang and Kauttonen, Janne and Zhao, Guoying},
  journal={IEEE Transactions on Affective Computing},
  volume={13},
  number={4},
  pages={2028--2046},
  year={2022},
  publisher={IEEE}
}

@article{he2025survey,
  title={A survey on uncertainty quantification methods for deep learning},
  author={He, Wenchong and Jiang, Zhe and Xiao, Tingsong and Xu, Zelin and Li, Yukun},
  journal={ACM Computing Surveys},
  year={2025},
  publisher={ACM New York, NY}
}

@article{ekman1978facial,
  title={Facial action coding system},
  author={Ekman, Paul and Friesen, Wallace V},
  journal={Environmental Psychology \& Nonverbal Behavior},
  year={1978}
}

@inproceedings{song2021uncertain,
  title={Uncertain graph neural networks for facial action unit detection},
  author={Song, Tengfei and Chen, Lisha and Zheng, Wenming and Ji, Qiang},
  booktitle={Proceedings of the AAAI Conference on Artificial Intelligence},
  volume={35},
  number={7},
  pages={5993--6001},
  year={2021}
}

@article{shao2021jaa,
  title={Jaa-net: joint facial action unit detection and face alignment via adaptive attention},
  author={Shao, Zhiwen and Liu, Zhilei and Cai, Jianfei and Ma, Lizhuang},
  journal={International Journal of Computer Vision},
  volume={129},
  number={2},
  pages={321--340},
  year={2021},
  publisher={Springer}
}

@article{tong2007facial,
  title={Facial action unit recognition by exploiting their dynamic and semantic relationships},
  author={Tong, Yan and Liao, Wenhui and Ji, Qiang},
  journal={IEEE transactions on pattern analysis and machine intelligence},
  volume={29},
  number={10},
  pages={1683--1699},
  year={2007},
  publisher={IEEE}
}

@inproceedings{niu2019local,
  title={Local relationship learning with person-specific shape regularization for facial action unit detection},
  author={Niu, Xuesong and Han, Hu and Yang, Songfan and Huang, Yan and Shan, Shiguang},
  booktitle={Proceedings of the IEEE/CVF Conference on computer vision and pattern recognition},
  pages={11917--11926},
  year={2019}
}

@article{shao2019facial,
  title={Facial action unit detection using attention and relation learning},
  author={Shao, Zhiwen and Liu, Zhilei and Cai, Jianfei and Wu, Yunsheng and Ma, Lizhuang},
  journal={IEEE transactions on affective computing},
  volume={13},
  number={3},
  pages={1274--1289},
  year={2019},
  publisher={IEEE}
}

@inproceedings{jacob2021facial,
  title={Facial action unit detection with transformers},
  author={Jacob, Geethu Miriam and Stenger, Bjorn},
  booktitle={Proceedings of the IEEE/CVF conference on computer vision and pattern recognition},
  pages={7680--7689},
  year={2021}
}

@inproceedings{yang2023fan,
  title={Fan-trans: Online knowledge distillation for facial action unit detection},
  author={Yang, Jing and Shen, Jie and Lin, Yiming and Hristov, Yordan and Pantic, Maja},
  booktitle={Proceedings of the IEEE/CVF Winter Conference on Applications of Computer Vision},
  pages={6019--6027},
  year={2023}
}

@inproceedings{ijcai2022p173,
  title     = {Learning Multi-dimensional Edge Feature-based AU Relation Graph for Facial Action Unit Recognition},
  author    = {Luo, Cheng and Song, Siyang and Xie, Weicheng and Shen, Linlin and Gunes, Hatice},
  booktitle = {Proceedings of the Thirty-First International Joint Conference on
               Artificial Intelligence, {IJCAI-22}},
  publisher = {International Joint Conferences on Artificial Intelligence Organization},
  editor    = {Lud De Raedt},
  pages     = {1239--1246},
  year      = {2022},
  month     = {7},
  note      = {Main Track}
}

@inproceedings{li2021integrating,
  title={Integrating semantic and temporal relationships in facial action unit detection},
  author={Li, Zhihua and Deng, Xiang and Li, Xiaotian and Yin, Lijun},
  booktitle={Proceedings of the 29th ACM international conference on multimedia},
  pages={5519--5527},
  year={2021}
}

@inproceedings{chang2020data,
  title={Data uncertainty learning in face recognition},
  author={Chang, Jie and Lan, Zhonghao and Cheng, Changmao and Wei, Yichen},
  booktitle={Proceedings of the IEEE/CVF conference on computer vision and pattern recognition},
  pages={5710--5719},
  year={2020}
}

@article{wang2021can,
  title={Can multi-label classification networks know what they don’t know?},
  author={Wang, Haoran and Liu, Weitang and Bocchieri, Alex and Li, Yixuan},
  journal={Advances in Neural Information Processing Systems},
  volume={34},
  pages={29074--29087},
  year={2021}
}

@article{li2018eac,
  title={Eac-net: Deep nets with enhancing and cropping for facial action unit detection},
  author={Li, Wei and Abtahi, Farnaz and Zhu, Zhigang and Yin, Lijun},
  journal={IEEE transactions on pattern analysis and machine intelligence},
  volume={40},
  number={11},
  pages={2583--2596},
  year={2018},
  publisher={IEEE}
}

@article{shao2026constrained,
  title={Constrained and directional ensemble attention for facial action unit detection},
  author={Shao, Zhiwen and Chen, Bikuan and Zhou, Yong and Shi, Xuehuai and Li, Canlin and Ma, Lizhuang and Yeung, Dit-Yan},
  journal={Pattern Recognition},
  volume={169},
  pages={111904},
  year={2026},
  publisher={Elsevier}
}

@article{zhang2014bp4d,
  title={Bp4d-spontaneous: a high-resolution spontaneous 3d dynamic facial expression database},
  author={Zhang, Xing and Yin, Lijun and Cohn, Jeffrey F and Canavan, Shaun and Reale, Michael and Horowitz, Andy and Liu, Peng and Girard, Jeffrey M},
  journal={Image and Vision Computing},
  volume={32},
  number={10},
  pages={692--706},
  year={2014},
  publisher={Elsevier}
}

@article{mavadati2013disfa,
  title={Disfa: A spontaneous facial action intensity database},
  author={Mavadati, S Mohammad and Mahoor, Mohammad H and Bartlett, Kevin and Trinh, Philip and Cohn, Jeffrey F},
  journal={IEEE Transactions on Affective Computing},
  volume={4},
  number={2},
  pages={151--160},
  year={2013},
  publisher={IEEE}
}

@inproceedings{zhang2018classifier,
  title={Classifier learning with prior probabilities for facial action unit recognition},
  author={Zhang, Yong and Dong, Weiming and Hu, Bao-Gang and Ji, Qiang},
  booktitle={Proceedings of the IEEE Conference on Computer Vision and Pattern Recognition},
  pages={5108--5116},
  year={2018}
}

@inproceedings{li2019semantic,
  title={Semantic relationships guided representation learning for facial action unit recognition},
  author={Li, Guanbin and Zhu, Xin and Zeng, Yirui and Wang, Qing and Lin, Liang},
  booktitle={Proceedings of the AAAI conference on artificial intelligence},
  volume={33},
  number={01},
  pages={8594--8601},
  year={2019}
}

@article{song2022heterogeneous,
  title={Heterogeneous spatio-temporal relation learning network for facial action unit detection},
  author={Song, Wenyu and Shi, Shuze and Dong, Yu and An, Gaoyun},
  journal={Pattern Recognition Letters},
  volume={164},
  pages={268--275},
  year={2022},
  publisher={Elsevier}
}

@inproceedings{li2023knowledge,
  title={Knowledge-spreader: Learning semi-supervised facial action dynamics by consistifying knowledge granularity},
  author={Li, Xiaotian and Zhang, Xiang and Wang, Taoyue and Yin, Lijun},
  booktitle={Proceedings of the IEEE/CVF international conference on computer vision},
  pages={20979--20989},
  year={2023}
}

@article{yang2023toward,
  title={Toward robust facial action units’ detection},
  author={Yang, Jing and Hristov, Yordan and Shen, Jie and Lin, Yiming and Pantic, Maja},
  journal={Proceedings of the IEEE},
  volume={111},
  number={10},
  pages={1198--1214},
  year={2023},
  publisher={IEEE}
}

@inproceedings{wang2024multi,
  title={Multi-scale dynamic and hierarchical relationship modeling for facial action units recognition},
  author={Wang, Zihan and Song, Siyang and Luo, Cheng and Deng, Songhe and Xie, Weicheng and Shen, Linlin},
  booktitle={Proceedings of the IEEE/CVF Conference on Computer Vision and Pattern Recognition},
  pages={1270--1280},
  year={2024}
}

@article{chang2024facial,
  title={Facial Action Unit Recognition Enhanced by Text Descriptions of FACS},
  author={Chang, Yanan and Zhang, Caichao and Wu, Yi and Wang, Shangfei},
  journal={IEEE Transactions on Affective Computing},
  year={2024},
  publisher={IEEE}
}

@article{liu2024multi,
  title={Multi-scale promoted self-adjusting correlation learning for facial action unit detection},
  author={Liu, Xin and Yuan, Kaishen and Niu, Xuesong and Shi, Jingang and Yu, Zitong and Yue, Huanjing and Yang, Jingyu},
  journal={IEEE Transactions on Affective Computing},
  year={2024},
  publisher={IEEE}
}

@article{shao2025facial,
  title={Facial action unit detection by adaptively constraining self-attention and causally deconfounding sample},
  author={Shao, Zhiwen and Zhu, Hancheng and Zhou, Yong and Xiang, Xiang and Liu, Bing and Yao, Rui and Ma, Lizhuang},
  journal={International Journal of Computer Vision},
  volume={133},
  number={4},
  pages={1711--1726},
  year={2025},
  publisher={Springer}
}

@inproceedings{kingma2013auto,
  author       = {Diederik P. Kingma and
                  Max Welling},
  title        = {Auto-Encoding Variational Bayes},
  booktitle    = {2nd International Conference on Learning Representations, {ICLR} 2014,
                  Banff, AB, Canada, April 14-16, 2014, Conference Track Proceedings},
  year         = {2014}}

@article{2017Graph,
  title={Graph Attention Networks},
  author={ Velikovi, Petar  and  Cucurull, Guillem  and  Casanova, Arantxa  and  Romero, Adriana  and Liò, Pietro and  Bengio, Yoshua },
  journal={CoRR},
  year={2017},
}

@article{zhu2025towards,
  title={Towards a robust group-level emotion recognition via uncertainty-aware learning},
  author={Zhu, Qing and Mao, Qirong and Zhang, Jialin and Huang, Xiaohua and Zheng, Wenming},
  journal={IEEE Transactions on Affective Computing},
  year={2025},
  publisher={IEEE}
}

@article{zhou2025ua,
  title={UA-FER: Uncertainty-aware representation learning for facial expression recognition},
  author={Zhou, Haoliang and Huang, Shucheng and Xu, Yuqiao},
  journal={Neurocomputing},
  volume={621},
  pages={129261},
  year={2025},
  publisher={Elsevier}
}

@inproceedings{chen2022uncertainty,
  title={Uncertainty-Aware Representation Learning for Action Segmentation.},
  author={Chen, Lei and Li, Muheng and Duan, Yueqi and Zhou, Jie and Lu, Jiwen},
  booktitle={IJCAI},
  volume={2},
  pages={6},
  year={2022}
}

@inproceedings{zhao2023open,
  title={Open set action recognition via multi-label evidential learning},
  author={Zhao, Chen and Du, Dawei and Hoogs, Anthony and Funk, Christopher},
  booktitle={Proceedings of the IEEE/CVF conference on computer vision and pattern recognition},
  pages={22982--22991},
  year={2023}
}

@article{kohl2018probabilistic,
  title={A probabilistic u-net for segmentation of ambiguous images},
  author={Kohl, Simon and Romera-Paredes, Bernardino and Meyer, Clemens and De Fauw, Jeffrey and Ledsam, Joseph R and Maier-Hein, Klaus and Eslami, SM and Jimenez Rezende, Danilo and Ronneberger, Olaf},
  journal={Advances in neural information processing systems},
  volume={31},
  year={2018}
}

@article{sohn2015learning,
  title={Learning structured output representation using deep conditional generative models},
  author={Sohn, Kihyuk and Lee, Honglak and Yan, Xinchen},
  journal={Advances in neural information processing systems},
  volume={28},
  year={2015}
}

@inproceedings{qin2022deep,
  title={Deep evidential learning with noisy correspondence for cross-modal retrieval},
  author={Qin, Yang and Peng, Dezhong and Peng, Xi and Wang, Xu and Hu, Peng},
  booktitle={Proceedings of the 30th ACM International Conference on Multimedia},
  pages={4948--4956},
  year={2022}
}

@article{sensoy2018evidential,
  title={Evidential deep learning to quantify classification uncertainty},
  author={Sensoy, Murat and Kaplan, Lance and Kandemir, Melih},
  journal={Advances in neural information processing systems},
  volume={31},
  year={2018}
}

@inproceedings{deng2009imagenet,
  title={Imagenet: A large-scale hierarchical image database},
  author={Deng, Jia and Dong, Wei and Socher, Richard and Li, Li-Jia and Li, Kai and Fei-Fei, Li},
  booktitle={2009 IEEE conference on computer vision and pattern recognition},
  pages={248--255},
  year={2009},
  organization={Ieee}
}

@article{SHANG2026112746,
title = {Learning contrastive feature representations for facial action unit detection},
journal = {Pattern Recognition},
volume = {173},
pages = {112746},
year = {2026},
issn = {0031-3203},
author = {Ziqiao Shang and Bin Liu and Fengmao Lv and Fei Teng and Tianrui Li and Lan-Zhe Guo}
}

@inproceedings{loshchilovdecoupled,
  title={Decoupled Weight Decay Regularization},
  author={Loshchilov, Ilya and Hutter, Frank},
  booktitle={International Conference on Learning Representations},
  year={2018}
}

@article{yan2022weakly,
  title={Weakly supervised regional and temporal learning for facial action unit recognition},
  author={Yan, Jingwei and Wang, Jingjing and Li, Qiang and Wang, Chunmao and Pu, Shiliang},
  journal={IEEE Transactions on Multimedia},
  volume={25},
  pages={1760--1772},
  year={2022},
  publisher={IEEE}
}

@INPROCEEDINGS{9710171,
  author={Ridnik, Tal and Ben-Baruch, Emanuel and Zamir, Nadav and Noy, Asaf and Friedman, Itamar and Protter, Matan and Zelnik-Manor, Lihi},
  booktitle={2021 IEEE/CVF International Conference on Computer Vision (ICCV)}, 
  title={Asymmetric Loss For Multi-Label Classification}, 
  year={2021},
  volume={},
  number={},
  pages={82-91}
  }

@inproceedings{li2021your,
  title={Your “attention” deserves attention: A self-diversified multi-channel attention for facial action analysis},
  author={Li, Xiaotian and Li, Zhihua and Yang, Huiyuan and Zhao, Geran and Yin, Lijun},
  booktitle={2021 16th IEEE International Conference on Automatic Face and Gesture Recognition (FG 2021)},
  pages={01--08},
  year={2021},
  organization={IEEE}
}

@INPROCEEDINGS{11209184,
  author={Xing, Bohao and Yuan, Kaishen and Yu, Zitong and Liu, Xin and Kälviäinen, Heikki},
  booktitle={2025 IEEE International Conference on Multimedia and Expo (ICME)}, 
  title={AU-TTT: Vision Test-Time Training model for Facial Action Unit Detection}, 
  year={2025},
  volume={},
  number={},
  pages={1-6}
}

@article{han2022trusted,
  title={Trusted multi-view classification with dynamic evidential fusion},
  author={Han, Zongbo and Zhang, Changqing and Fu, Huazhu and Zhou, Joey Tianyi},
  journal={IEEE transactions on pattern analysis and machine intelligence},
  volume={45},
  number={2},
  pages={2551--2566},
  year={2022},
  publisher={IEEE}
}

@article{adyapady2023comprehensive,
  title={A comprehensive review of facial expression recognition techniques},
  author={Adyapady, R Rashmi and Annappa, B},
  journal={Multimedia Systems},
  volume={29},
  number={1},
  pages={73--103},
  year={2023},
  publisher={Springer}
}
\bibliographystyle{ACM-Reference-Format}

\appendix

\end{document}